\documentclass[11pt]{article}
\usepackage[margin=1in]{geometry}
\usepackage{amsmath,amssymb}
\usepackage{graphicx}
\usepackage{booktabs}
\usepackage{microtype}
\usepackage{placeins}
\usepackage[numbers,sort&compress]{natbib}
\usepackage[colorlinks=true,linkcolor=blue,citecolor=blue,urlcolor=blue]{hyperref}
\graphicspath{{figures/}}

\newcommand{\Tf}{\ensuremath{T_F}}
\newcommand{\Tv}{\ensuremath{T_V}}
\newcommand{\phat}{\ensuremath{\hat{p}}}
\newcommand{\Rhat}{\ensuremath{\hat{R}}}
\newcommand{\tauthr}{\ensuremath{\tau}}

\title{It's the Problem, Not the Path:\\
Budget and Difficulty Confounds in LLM Reasoning Trajectories}
\author{Yigit Utku Bulut\thanks{Work conducted independently.}\\
Johannes Kepler University Linz\\
\texttt{yigit.utku.bulut@outlook.com}\\
\texttt{k12350740@students.jku.at}}
\date{\today}

\begin{document}
\maketitle

\begin{abstract}
Reasoning traces of large language models are widely read as containing
``breakthrough'' moments and early-legible fates. Both readings rest on
measurements missing a counterfactual control at the level of the claim;
we supply both controls. First, a restart-controlled truncation probe
separates \emph{when a solution fits the continuation budget} from
\emph{when a prefix carries value that fresh computation cannot buy},
comparing per-anchor continuation solve rates against from-scratch
restart curves at matched total generated-token budget. Applied to 178
problem--model cells (89 MATH problems $\times$ two small open models, an
outcome-blind but difficulty-targeted cohort), exactly 1 of 178 cells
survives as prefix-limited; restart dose--response separates a
compute-starved model from a capability-limited one; and wherever the
matched budget lies inside the restart grid, continuing the model's own
prefix beats restarting (9 of 9) --- predominantly compute compression
rather than expanded reachability. Second, a pre-registered,
difficulty-controlled test finds no detectable outcome information in
early-window internal signals beyond a problem-difficulty baseline, and
two generation-free analyses of public corpora show why this control is
needed: a trace-blind difficulty proxy reaches AUROC 0.873 on 192K
DeepSeek-R1 generations --- inside the published probe range --- and
a closely matched reconstruction of the closest published early-window
positive recovers a comparable pooled result (0.849) while
within problem it is statistically indistinguishable from chance at all
ten anchors (0.496 at $t{=}4$); a post-hoc within-targeted probe finds
only a small average residual, concentrated in three low-failure
problems. High pooled
probe AUROCs cannot by
themselves establish within-attempt information; a question-only baseline
or within-problem evaluation is required.
\end{abstract}

\section{Introduction}\label{sec:intro}

Two appealing beliefs circulate about the reasoning traces of large
language models. The first is that traces contain \emph{breakthrough
moments} --- points where accumulated reasoning suddenly unlocks a
solution, popularized as ``aha moments'' in reinforcement-trained
reasoners \citep{deepseekr1}. The second is that a run's fate is legible
early: probes on hidden states reportedly predict correctness
with AUROC $0.79$--$0.95$, sometimes from the first four tokens
\citep{zhang2025knowright, yuan2026diagnostic, david2025temporal}, and
serving systems already allocate compute on such signals
\citep{fu2024certaindex, fu2025deepconf}. Both beliefs, if true, would
matter: the first for what test-time compute buys, the second for
adaptive inference and interpretability.

Both beliefs rest on measurements with a missing control, and the two
missing controls mirror each other: a claim about a trajectory requires a
counterfactual control at the level of the claim --- what this
computation buys over fresh computation, what this attempt reveals beyond
its problem. Breakthrough claims rest on
truncate-and-resample probes: cut the trace at position $t$, sample
continuations under a budget, and call the first high-solve-rate anchor a
breakthrough. But a crossing of that kind conflates two events ---
\emph{the prefix accumulated value} and \emph{the remaining work began to
fit the budget}. Distinguishing them requires a from-scratch \emph{restart}
control at matched total generated-token budget; we found no prior work
combining per-anchor continuation estimates with a per-problem restart
curve at matched budget (Section~\ref{sec:related}).
Prediction claims rest on probes evaluated across problems, where a probe
can score well by reading \emph{which problem it is} --- its difficulty
--- rather than how the attempt is going. Distinguishing those requires a
question-only baseline and a within-problem evaluation, which the
published positives do not report.

This paper supplies both controls and reports what they change. We build a
restart-controlled, budget-indexed, censoring-aware truncation probe
(Section~\ref{sec:method}): per-anchor continuation solve rates
$\phat(t)$ are compared against restart curves $R(C)$ at matched total
generated-token budget, splitting the naive breakthrough time into a budget-fit time
\Tf{} and a prefix-value time \Tv, with attempt-level noise hardened by
replication and pooling. We apply it to 178 problem--model cells (89 MATH
problems $\times$ two small open models, 16K-token instrumented traces),
and we complement it with two analyses of public reasoning-model corpora
at larger scale.

The instrument returns a consistent picture:

\begin{itemize}
\item \textbf{Breakthroughs are mostly budget artifacts.} Exactly 1 of 178
cells survives as prefix-limited (three cross the advantage margin; an
in-grid restart already solves two); 98 cells collected after the
rules were frozen contributed zero new cases
(Section~\ref{sec:results-regimes}).
\item \textbf{Restart dose--response separates failure modes.} One model's
unsolved problems dissolve as budget grows ($0\%\to79\%$,
compute-starved); the other's do not budge (capability-limited within
the measured restart-budget range).
\item \textbf{Accumulated reasoning is predominantly compute
compression.} At matched total token budget the model's own prefix beats
restarting wherever the comparison is exactly matched (9/9, with 2 wins
and 2 ties on boundary proxies), yet a larger restart budget reaches the
same threshold in 11/13 and the prefix's own rate in 9/13: within the
measured budget range, long reasoning mostly buys the same successes
cheaper rather than reaching otherwise unreachable ones.
\item \textbf{Solvability is not binary.} A third of ambiguous
intermediate states remain intermediate after eight attempts (success
$0.3$--$0.6$), a band reproduced in the independent second half of
attempts.
\item \textbf{Early internal signals carry no detectable outcome
information beyond difficulty.} A pre-registered, single-shot test on a
held-out split finds no detectable gain at the frozen forecast point, and
a labeled post-hoc sweep finds none at any window from 128 to 2,048
tokens (Section~\ref{sec:results-prediction}); on public data, a
trace-blind difficulty proxy reaches 0.873 on 192K DeepSeek-R1
generations --- inside the published probe range --- and
a closely matched reconstruction of the closest published early-window
positive recovers a comparable pooled result at $t{=}4$ (0.849)
while within problem it is statistically indistinguishable from chance at
all ten anchors (0.496 at $t{=}4$); a post-hoc within-targeted check
finds only a small average residual, concentrated in three low-failure
problems.
\end{itemize}

The claims are scoped deliberately. We study two small open-weight models
(one an explicitly reasoning-tuned release, both run in thinking mode)
plus public corpora from the R1 family; we do not claim internal states
never carry within-attempt information --- at intermediate-answer
positions, in larger models, or in other domains the verdict may differ. What we do claim is that the two controls change
conclusions wherever we have applied them, that they are cheap (our public
data analyses required no generation at all), and that trajectory-level
claims about reasoning should be reported with them. Every
decision rule in this study was frozen in committed amendments before the
outcomes it governs were observed, gate failures included
(Section~\ref{sec:method}, App.~\ref{app:amendments}). Code, the frozen
amendments, and all result artifacts are available at
\url{https://github.com/bulutyigit/problem-not-path}.

\section{A Restart-Controlled Probe of Reasoning Value}\label{sec:method}

\subsection{Setup}

We study two small open-weight models with distinct architectures and
training pipelines --- Gemma-4 E4B and Ministral-3 3B (an explicitly
reasoning-tuned release)\footnote{\raggedright Checkpoints
\texttt{google/gemma-4-E4B-it} and
\texttt{mistralai/Ministral-3-3B-Reasoning-2512}, run as their
\texttt{mlx-community} 4-bit conversions.\par} --- run locally in thinking
mode, in 4-bit MLX quantization, with full
token-level instrumentation (per-token logits and final-layer hidden
states). Problems come from the MATH benchmark \citep{hendrycks2021math}.
Because breakthrough measurement is only informative on problems a model
sometimes solves, cohorts were selected by outcome-blind
intermediate-difficulty screens, frozen before any probe outcome existed: a
development cohort (20 problems), a supplement cohort (20), and an expansion
cohort (49) whose screen was repaired after a recorded gate failure
(App.~\ref{app:amendments}). Together they yield 89 problems and $89 \times
2 = 178$ problem--model cells. Each cell has one instrumented base
trajectory generated with a 16{,}384-token budget. Research splits (54
train / 18 validation / 17 test problems) were assigned before any outcome
was observed; the test split is spent exactly once, in
Section~\ref{sec:results-prediction}.

\subsection{Truncation probe and the budget-fit time \Tf}

The probe follows the truncation-and-resampling tradition
\citep{lanham2023,bogdan2025anchors}: truncate the model's \emph{own}
trajectory at an anchor $t$ (log-spaced from 16 to 8{,}192 tokens), and
sample $m=4$ continuations from the truncated prefix under a fixed budget
of $B = 1{,}024$ reasoning tokens plus a 512-token answer reserve, with
deterministic per-branch seeds. The per-anchor solve rate
$\phat(t; B)$ estimates the value of the partial state. A
\emph{breakthrough} is the first anchor with $\phat \ge \tauthr$
($\tauthr = 0.75$) that remains above threshold at the next anchor;
crossings are refined by bisection and recorded as intervals, and
trajectories whose curve never crosses are right-censored at their final
anchor. We call the resulting time \Tf$(B)$, the \emph{budget-fit time}:
the point at which a solution first fits the continuation budget. As
Section~\ref{sec:results-regimes} shows, \Tf{} is what an uncontrolled
probe reports as a breakthrough --- and it conflates two very different
things.

\subsection{Restart control and the prefix-value time \Tv}

To separate ``the prefix carries value'' from ``the budget became
sufficient,'' we measure each problem's \emph{restart curve} $R(C)$: the
solve rate of from-scratch attempts (empty prefix, same prompt) at budgets
$C \in \{1{,}024, 2{,}048, 4{,}096, 8{,}192\}$, four attempts each, with
$\Rhat$ interpolated log-linearly between grid points. Budget matching
throughout equates \emph{generated tokens}; FLOPs, latency, and KV-cache
reuse are not measured (Section~\ref{sec:discussion}). The
\emph{advantage} of a prefix at matched total generated-token budget is
$\mathrm{adv}(t) = \phat(t; B) - \Rhat(t + B)$, and the prefix-value time
$\Tv(\delta)$ is the earliest stable anchor with $\phat \ge \tauthr$ and
$\mathrm{adv} \ge \delta$ (primary margin $\delta = 0.5$; $0.25$ and
$0.75$ reported as sensitivity, along with a conservative variant that
takes the upper envelope of the bracketing grid values of $\Rhat$). Cells
are then classified by frozen precedence: \emph{instant} (event interval
upper bound $\le 16$ tokens), \emph{budget-limited} ($\Rhat(4{,}096) \ge
\tauthr$: a restart solves it), \emph{prefix-limited} (a $\delta = 0.5$
crossing exists), \emph{no-crossing} (\Tf{} exists but no advantage
crossing), \emph{terminal} (a replicated event at the final anchor with no
stability anchor; annotated, never promoted to \Tv), and \emph{unsolved}.

\subsection{Noise hardening}

With $m = 4$, a $3/4$ crossing occurs with probability $0.31$ under a true
success rate of $0.5$, so single-shot labels are optimistic. Two frozen
rules harden them. First, threshold-censored trajectories (final-anchor
rate $\ge \tauthr$, no stability anchor available) get four additional
branches, and a terminal event is recorded only if the pooled rate reaches
$\ge 6/8$ (under $p = 0.5$ this has probability $0.14$). Second, every
ambiguous cell ($1$--$3$ successes of $4$) across the expansion cohort was
enlarged to eight attempts before labels were derived (148 cells). Both
rules cut in both directions in practice: replication confirmed 13 of 16
candidates for one model and rejected the other model's single candidate,
and pooling removed six events while adding one.

\subsection{Pre-registration discipline}

Every decision rule above was frozen in written, committed amendments
before the outcomes it governs were observed; gate failures (two cohort
gates and one pilot gate) are recorded as failures and resolved by
amendment, never by relabeling. The confirmatory prediction protocol of
Section~\ref{sec:results-prediction} --- endpoints, feature sets, model
class, power gates, and success criterion --- was committed before the
single test-split evaluation, and every post-hoc analysis is labeled as
such in the amendment that reports it. Appendix~\ref{app:amendments} gives
the full timeline.

\section{Results I: Breakthroughs Under Control}\label{sec:results-regimes}

\subsection{Most measured breakthroughs are budget artifacts}

\begin{figure}[t]\centering
\includegraphics[width=\linewidth]{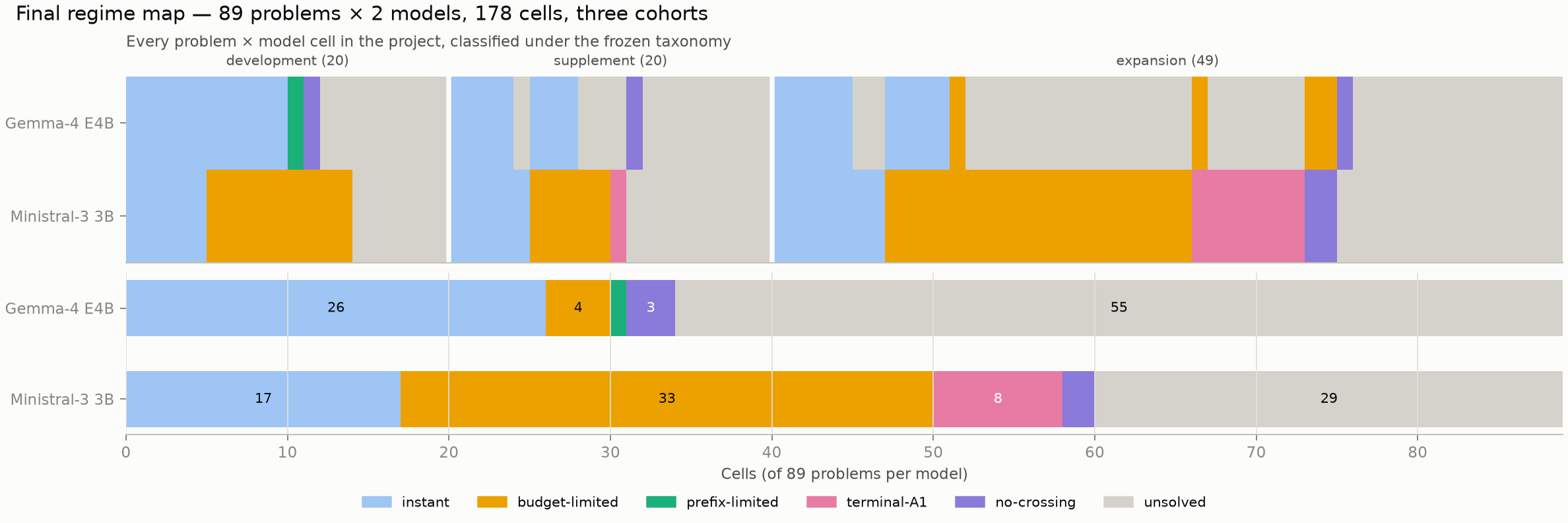}
\caption{The final regime map: 89 problems $\times$ 2 models classified
under the frozen rules (top: per-cell strip grouped by cohort; bottom:
per-model totals). One cell of 178 is prefix-limited.}
\label{fig:regimemap}\end{figure}

The restart control changes what the probe data mean. In the development
cohort, Ministral-3 exhibited apparently clean mid-trajectory crossings ---
until a budget-sensitivity re-probe showed the same problems solved from
16-token prefixes once the continuation budget was raised to 4{,}096
tokens: the ``breakthroughs'' marked where solutions began to fit the
budget, not where the prefix accumulated value. Applied to all 178 cells,
the frozen taxonomy yields the map in Figure~\ref{fig:regimemap}: for
Ministral-3, 33 cells are budget-limited, 17 instant, 8 terminal, 2
no-crossing, and 29 unsolved; for Gemma-4, 26 instant, 4 budget-limited,
3 no-crossing, 55 unsolved --- and exactly \emph{one} prefix-limited cell
in the entire dataset (a development problem with $\Tv = 896$ and advantage
$0.75$). Decomposed: the advantage crossing itself occurs in three cells at
the primary margin; the frozen precedence classifies the other two as
budget-limited because a 4{,}096-token restart already solves them, and
four \emph{instant} cells additionally show an advantage at their
$\le 16$-token anchors. The 98 expansion cells, collected after the rules were frozen,
contributed zero new prefix-limited cases. Loosening the margin to
$\delta = 0.25$ adds four crossings, tightening to $0.75$ keeps two, and
the conservative $\Rhat$ envelope agrees with the primary labels on all
178 cells. Genuine prefix-locked value, at this scale and in these models,
is rare: 1 of 178 cells --- a descriptive count in an outcome-blind but
difficulty-targeted cohort with one base trajectory per cell (the 89
problems repeat across the two models, so cells are not independent and
no population-prevalence interval is implied).

\subsection{Restart dose--response separates failure modes}

\begin{figure}[t]\centering
\includegraphics[width=\linewidth]{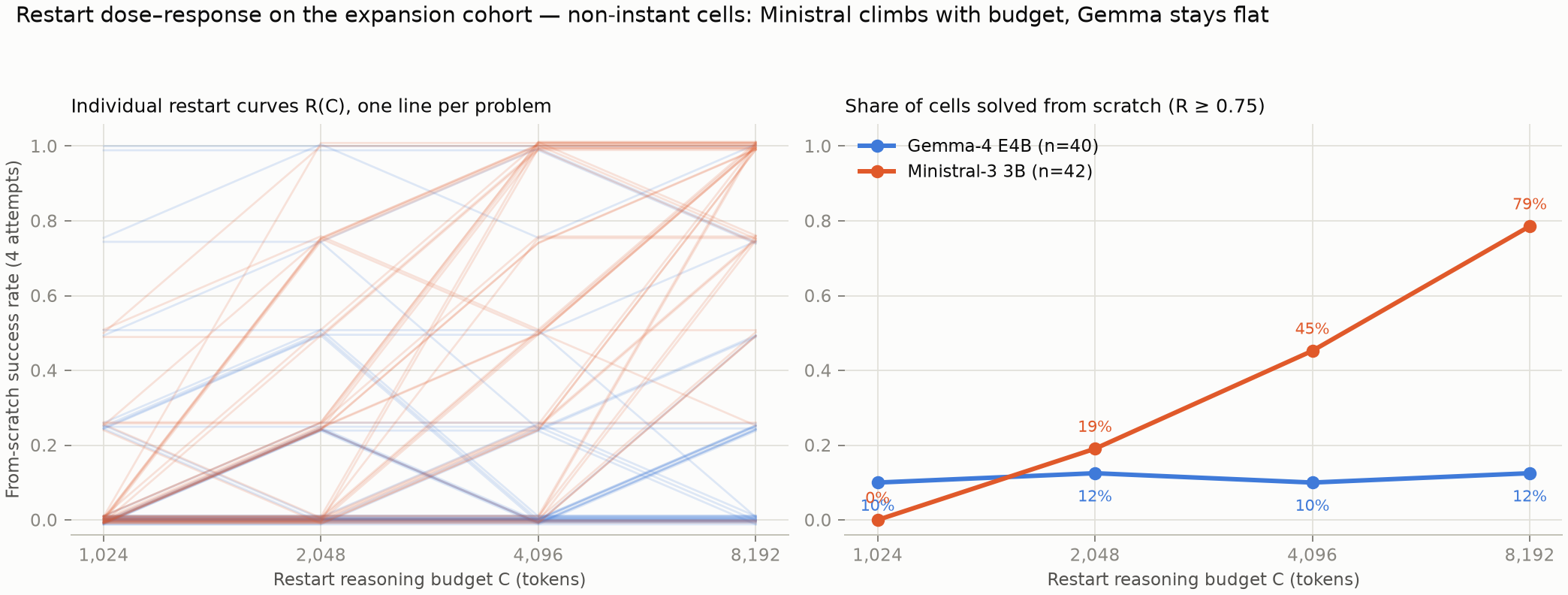}
\caption{Restart dose--response on non-instant expansion cells.
Individual curves (left) and share solved from scratch (right):
Ministral-3 climbs $0\%\to79\%$; Gemma-4 stays flat.}
\label{fig:dose}\end{figure}

The restart curves themselves carry the cleanest model-level finding
(Figure~\ref{fig:dose}). On expansion cells that are not instant, the
share of problems Ministral-3 solves from scratch climbs with budget ---
$0\% \to 19\% \to 45\% \to 79\%$ across $1{,}024 \to 8{,}192$ tokens ---
while Gemma-4 stays flat at $10$--$12\%$. Individual four-attempt curves
are noisy: 8 of Ministral-3's 42 show at least one decrease between
consecutive budgets (all of a single attempt, $0.25$; two cells revert
across the $\tauthr$ threshold), against 18 of Gemma-4's 40, five with
drops of at least $0.5$ and one threshold reversal --- larger-budget
restarts \emph{losing} solved problems, overthinking on natural problems
that complements constructed-task inverse scaling
\citep{gema2025inverse}. One model's failures are
\emph{compute-starved}; the other's are \emph{capability-limited} within
the measured restart-budget range. The two
models agree on a collapsed regime class for only 42 of 89 problems:
small models do not share a single ``small-model'' failure mode, which
also cautions against averaging such models in evaluations.

\subsection{Accumulated reasoning is predominantly compute compression}

\begin{figure}[t]\centering
\includegraphics[width=.9\linewidth]{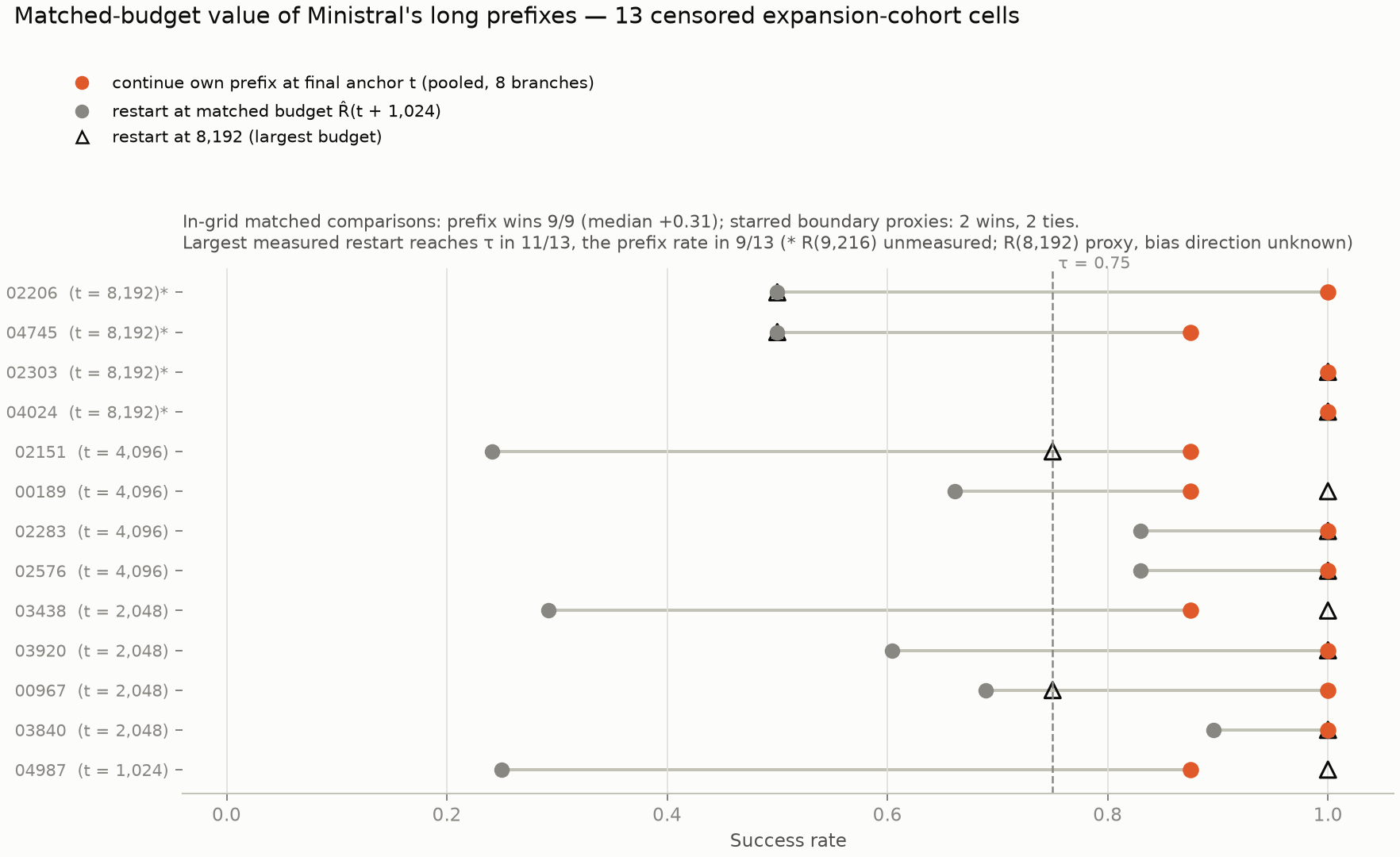}
\caption{Matched-budget value of long prefixes on 13 threshold-censored
cells: the prefix wins all nine exactly matched comparisons (starred
boundary proxies: two wins, two ties); the largest
measured restart reaches $\tauthr$ in 11/13 and the prefix's own rate in
9/13.}
\label{fig:matched}\end{figure}

The 13 threshold-censored Ministral-3 expansion cells --- trajectories
whose crossing was replicated only at the final anchor; under the frozen
precedence seven carry the terminal label, five budget-limited, and one
no-crossing --- allow the
sharpest statement of what a long prefix is worth
(Figure~\ref{fig:matched}). In the nine cells whose matched budget $t + B$ lies inside the restart
grid, continuing the model's own prefix beats restarting in all nine
(median advantage $+0.31$). The four cells anchored at $t = 8{,}192$
require $\Rhat(9{,}216)$, which was not measured; substituting
$\Rhat(8{,}192)$ makes them proxy comparisons whose bias direction is
unknown --- restart curves are not uniformly monotone --- and on these
proxies the prefix wins two and ties two. The largest measured restart
reaches $\tauthr$ in 11 of 13 cells and matches the prefix's own pooled
success rate in 9 of 13.
Accumulated reasoning is therefore predominantly consistent with
\emph{compression} --- reaching the same success threshold with fewer
total tokens --- rather than with expanded reachability, within the
measured budget range: in the two cells where no measured restart reaches
$\tauthr$ the question stays open at larger budgets, and binary
correctness compares success rates, not solution content. This
instantiates, empirically and per problem, the no-benefit branch of
the dichotomy of \citet{wolf2026theory} --- absent self-reflection that
reliably localizes early errors, conditioning on past attempts offers no
asymptotic benefit over independent restarts (their analysis concerns
attempt-conditioned search; our truncated-prefix continuation is the
analogous object) --- and it
gives the sequential-versus-parallel debate \citep{snell2024,
ghosal2025mirage, sharma2025sequentialedge} a quantity it has lacked: the
value of a partial trace expressed in fresh-token units.

\subsection{A persistent band of intermediate solvability}

\begin{figure}[t]\centering
\includegraphics[width=\linewidth]{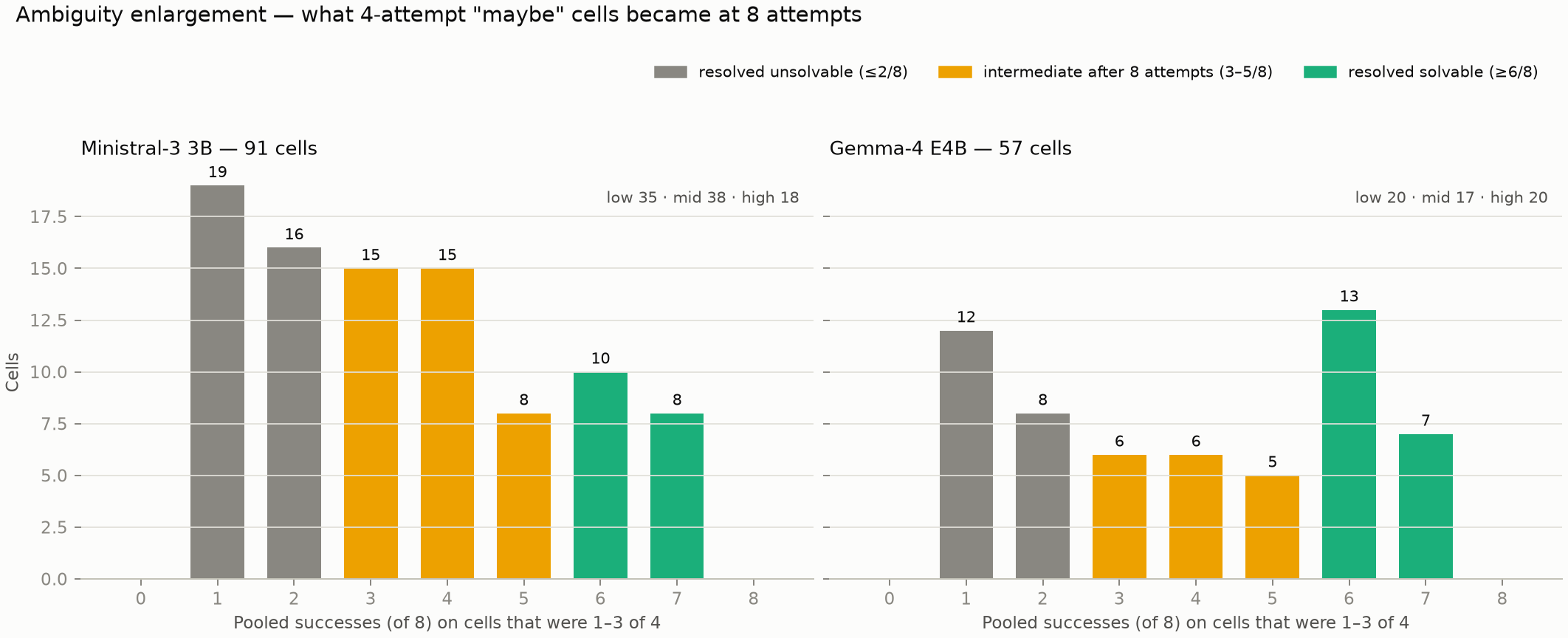}
\caption{What ambiguous $1$--$3/4$ cells became at eight attempts: about a
third remain intermediate, a pattern the independent second half of
attempts reproduces.}
\label{fig:band}\end{figure}

Enlarging all 148 ambiguous cells to eight attempts resolved 55 as
effectively unsolvable ($\le 2/8$) and 38 as solvable ($\ge 6/8$) --- and
left \emph{55 intermediate after eight attempts} ($3$--$5$ of $8$;
Figure~\ref{fig:band}). Because these cells were selected on their first
four attempts, pooled rates partly reflect selection; the independent
second half of attempts, however, reproduces the band --- 58\% of
enlarged cells land at $1$--$3$ of $4$ on branches never used for
selection --- so intermediate success probabilities are not a pure
selection artifact, though eight-attempt intervals remain wide. A sizable
band of intermediate-solvability states is the natural reading, which
undermines the single-crossing, monotone-value
picture implicit in binary-search process-labeling schemes
\citep{luo2024omegaprm} and in snowball-error accounts of monotone
degradation \citep{gan2025snowball}, and which converges with
trajectory-level evidence that recoverable and structural failures are
distinct populations \citep{islah2026}.

\section{Results II: A Difficulty-Controlled Null for Early Signals}\label{sec:results-prediction}

\subsection{A pre-registered, difficulty-controlled null}

Can the first tokens of a run predict its outcome beyond what the problem
itself predicts? We froze the answer procedure before evaluating it:
endpoints (primary: eventual success of the 16K run; secondary:
scratch-solvability at 4{,}096, i.e.\ $\Rhat(4{,}096) \ge \tauthr$ on
non-instant cells), feature sets (a question-only baseline versus the same baseline plus
fifteen frozen early-window dynamics summaries --- entropy, surprisal,
successive divergences, hidden-state geometry, spectral summaries over the
first 512 tokens; defined in Appendix~\ref{app:signals}), a logistic model, problem-grouped cross-validation on
train+validation, per-class power gates, and the success criterion: the
paired $\Delta$AUROC's problem-clustered 95\% CI must exclude zero. The
question-only baseline contains no signal from the run: the frozen
difficulty columns are the benchmark's human difficulty level, the topic
category, and five surface statistics of the problem text (character
count, whitespace-token count, and counts of numeric, operator, and
equation tokens), plus the model's identity --- everything knowable before
the first token is generated, at text level (a pre-generation
\emph{activation} probe would be a stronger baseline still
\citep{lugoloobi2026}; see Limitations). The test split was then evaluated
once.

\begin{figure}[t]\centering
\includegraphics[width=\linewidth]{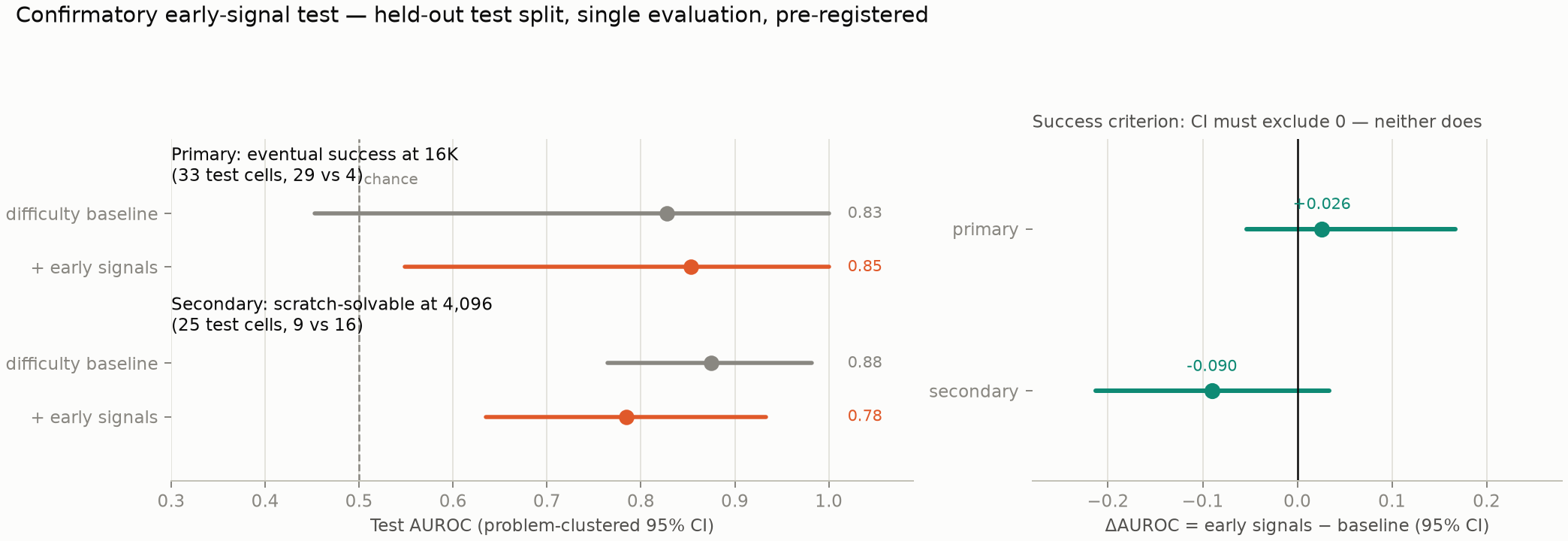}
\caption{The confirmatory test: single evaluation on the held-out split.
Neither endpoint's $\Delta$AUROC CI excludes zero.}
\label{fig:conf}\end{figure}

\begin{figure}[t]\centering
\includegraphics[width=\linewidth]{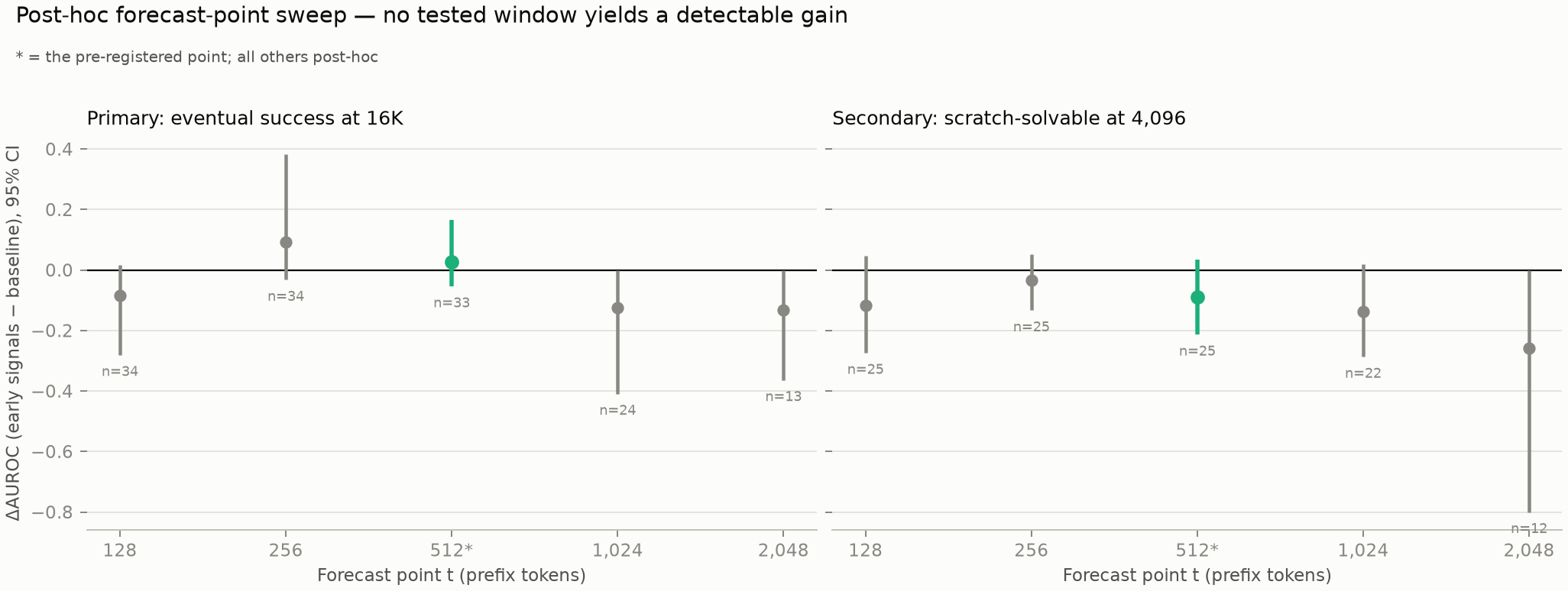}
\caption{Post-hoc forecast-point sweep: no tested window yields a
detectable gain.}
\label{fig:sweep}\end{figure}

Neither endpoint met the criterion (Figure~\ref{fig:conf}). Primary:
baseline AUROC $0.83$ versus $0.85$ with early signals, $\Delta = +0.026$
$[-0.054, +0.167]$. Secondary: baseline $0.88$ versus $0.78$,
$\Delta = -0.090$ $[-0.213, +0.033]$ --- the point estimate is negative,
consistent with uninformative added features. A level-free
sensitivity variant agrees, and the within-trajectory timing endpoint
remained below its power gate (four test-split interior events) and was
not fit. A post-hoc sweep of the forecast point over
$t \in \{128, 256, 1{,}024, 2{,}048\}$ --- labeled post-hoc in the
amendment --- found 8 of 10 point estimates negative and every CI
straddling zero (Figure~\ref{fig:sweep}): no tested alternative window
yields a detectable gain, though these post-hoc intervals are individually
underpowered. These are absences of detected gain,
not demonstrations of equivalence --- the primary CI's upper bound
($+0.167$) does not exclude moderate effects, and no equivalence margin
was pre-registered anywhere; what tighter intervals can support is an
explicit bound on effect size, as in the within-problem dissection
below.

Why the null has this shape is visible in the per-model split: the
difficulty baseline alone is near-perfect for Gemma-4 (test AUROC $1.0$
on both endpoints, $n{=}16$ and $11$) and mediocre for Ministral-3
($0.69$--$0.73$). Descriptively, difficulty leaves little to explain for
the capability-limited model, while the
budget-elastic model's outcomes carry residual stochastic variance --- the
band of Section~\ref{sec:results-regimes} --- that neither difficulty nor
early dynamics explain.

\subsection{Why pooled evaluation cannot see within-attempt information}

The confound has a simple formal core. Fix a problem--model cell $c$ and
treat attempts as conditionally i.i.d.\ given the cell: attempt $j$ has
outcome $Y_{c,j} \in \{0,1\}$ with $\Pr(Y_{c,j} = 1 \mid c) = p_c$. Call
a predictor \emph{question-only} if its score depends on the cell alone,
$S_{c,j} = f(c)$; the idealized case is $f(c) = p_c$. Its feasible
surrogate, the leave-one-out pass rate
$\hat d_{c,j} = \frac{1}{k_c - 1} \sum_{i \ne j} Y_{c,i}$, estimates
$p_c$ without using attempt $j$'s own outcome, but is \emph{not} itself
question-only --- it varies across attempts within a cell, with the
mechanical consequence recorded below. AUROC is a probability over discordant pairs,
$\mathrm{AUROC}(S) = \Pr(S_a > S_b \mid Y_a = 1, Y_b = 0) + \tfrac{1}{2}
\Pr(S_a = S_b \mid \cdot)$, and splitting the pairs by whether they share
a cell gives
\begin{equation}
\mathrm{AUROC}_{\mathrm{pooled}}(S) \;=\; \lambda \,
\mathrm{AUROC}_{\mathrm{within}}(S) \;+\; (1 - \lambda)\,
\mathrm{AUROC}_{\mathrm{between}}(S),
\label{eq:decomp}
\end{equation}
where $\lambda$ is the fraction of discordant pairs drawn from the same
cell. Two consequences follow. First, every question-only predictor has
$\mathrm{AUROC}_{\mathrm{within}} = \tfrac{1}{2}$ exactly (its score is
constant within a cell, so all within-cell pairs are ties): within-cell
evaluation isolates precisely the information that a difficulty
measurement cannot carry. Second, in the standard design with one attempt
per problem ($k_c = 1$), $\lambda = 0$ identically --- a pooled AUROC then
evaluates only between-cell ranking, a task for which ranking by $p_c$ is
optimal among question-only scores under conditional i.i.d.\ sampling,
and therefore cannot, by itself, attribute its value to within-attempt
information, however high it reads --- difficulty alone can produce any of
the reported numbers. The closest published early-window positive
\citep{david2025temporal} is of this form (one trajectory per problem). The two
designs in this section are the two escapes Eq.~\eqref{eq:decomp} allows:
with $k_c = 1$ the within component is inaccessible and the only recourse
is \emph{incremental} value over an explicit question-only baseline (our
confirmatory test above); with $k_c \gg 1$ the within component becomes
directly estimable (the dissection below). One bookkeeping fact used
there: within a cell with $S_c = \sum_i Y_{c,i}$ successes, the LOO score
takes exactly two values, $(S_c - 1)/(k_c - 1)$ for a correct attempt and
$S_c/(k_c - 1)$ for an incorrect one, ranking every incorrect attempt
above every correct one --- its within-AUROC is mechanically $0$, and the
neutral reference for ``no within-attempt information'' is $\tfrac{1}{2}$.

\subsection{The ceiling exists in the literature's own regime}

\begin{figure}[t]\centering
\includegraphics[width=.9\linewidth]{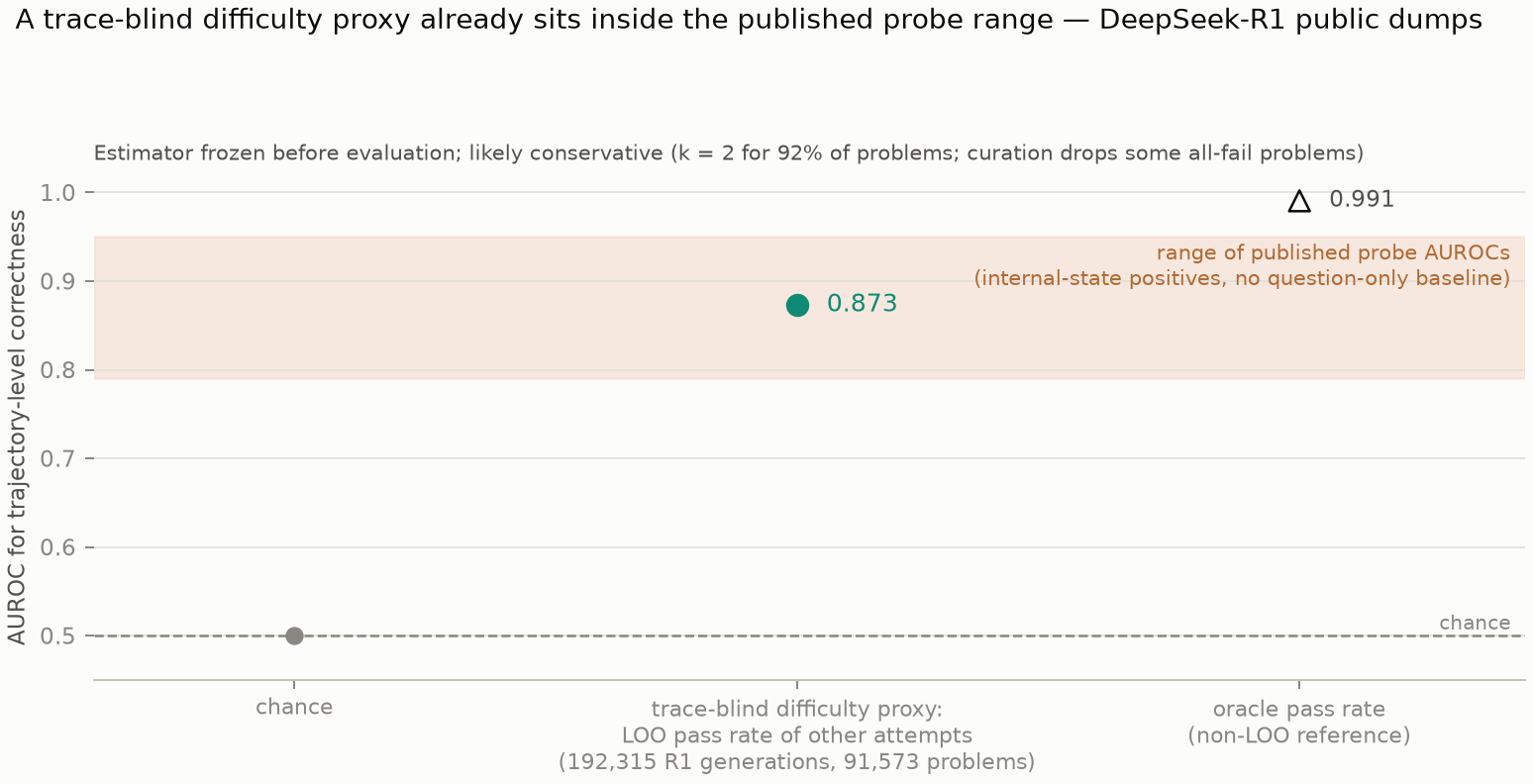}
\caption{A trace-blind difficulty proxy (the leave-one-out pass rate of
the problem's other attempts) reaches AUROC 0.873 on public DeepSeek-R1
generations --- inside the published probe range.}
\label{fig:ceiling}\end{figure}

Our cohort is small, so we asked the same question of public data at
scale. On 192{,}315 DeepSeek-R1 generations over 91{,}573 problems
\citep{openr1math}, a trace-blind difficulty proxy --- the leave-one-out
pass rate of the problem's \emph{other} attempts, for 92\% of problems a
single binary observation, reading neither the trace nor the question ---
achieves AUROC $0.873$ $[0.870, 0.876]$
(problem-clustered bootstrap; estimator frozen before evaluation, and
likely conservative, since dataset curation truncates the
difficulty range). That places a predictor that never reads the trace
squarely inside the $0.79$--$0.95$ range that internal-state probe papers report
\citep{zhang2025knowright, yuan2026diagnostic, sun2026trajectories,
david2025temporal} --- none of which report a question-only baseline
(Figure~\ref{fig:ceiling}; the partial exception,
\citet{lugoloobi2026}, baselines \emph{pre-generation} probes against
question-text features but never probes the reasoning window). Nor is the ceiling a mathematics
artifact: on 16{,}384 public GPQA-diamond \citep{rein2023gpqa} samples (64 questions $\times$
256 attempts, Llama-3.3-70B \citep{singhi2025solveverify}), the same
estimator reaches $0.917$ $[0.875, 0.938]$ --- despite a four-way
multiple-choice guessing floor that, under independent uniform guessing,
attenuates rather than inflates it.

\subsection{Dissecting a published probe positive in the reasoning-model
regime}

Finally, we reconstructed and then dissected the closest published
early-window setup in a public reasoning-model regime, using public data
and no generation: 128 MATH
problems with 256 R1-Distill-Qwen-7B samples each
\citep{singhi2025math128dump} --- a dump released by the first author of
\citet{singhi2025solveverify} alongside that paper's data, which the
paper itself does not describe --- re-scored by
teacher-forcing the dump's own tokens through the 4-bit model (top-1
fidelity $0.906$; protocol frozen before extraction). The 256-per-problem
design permits the control the published early-window positives do not
report: evaluating the
\emph{same} probe within problem, where difficulty is held fixed by
construction.

\begin{figure}[t]\centering
\includegraphics[width=\linewidth]{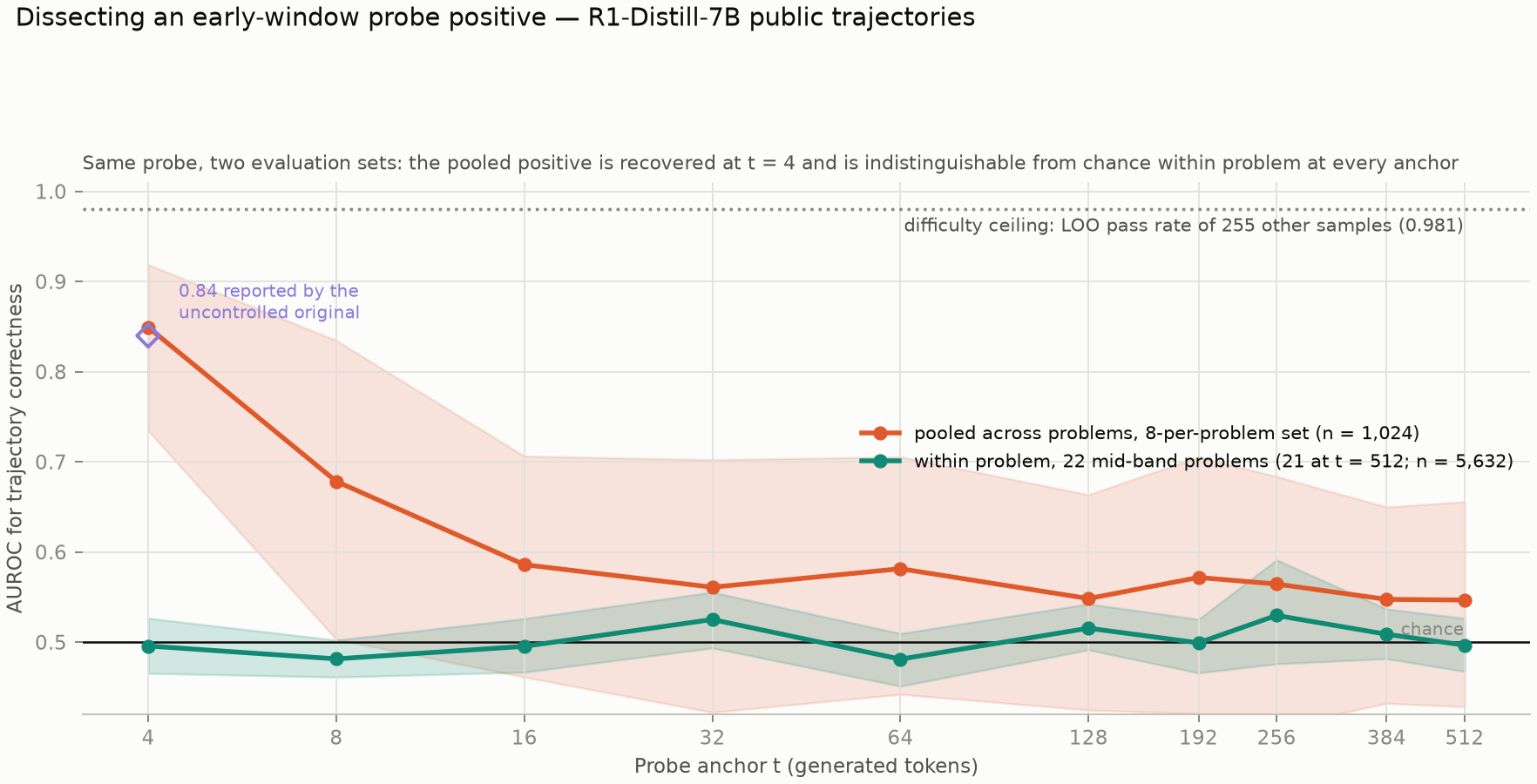}
\caption{Same probe, two evaluation sets, 6{,}480 public R1-Distill-7B
trajectories: the pooled positive is recovered at $t=4$ and is
indistinguishable from chance within problem at every anchor.}
\label{fig:dissect}\end{figure}

The uncontrolled positive is recovered: a last-four-token hidden-state probe
at $t = 4$ reaches pooled AUROC $0.849$ $[0.735, 0.919]$ on the
8-per-problem evaluation set ($n{=}1{,}024$), matching the $0.84$
reported by \citet{david2025temporal}. The pooled reconstruction and the
within-problem analysis use different evaluation sets: the former
uses the frozen 8-per-problem pooled subset, the latter all 256
samples of the 22 problems with sufficient outcome variation --- problems
that hold 94\% of the dump's failures and 84\% of its same-cell
discordant pairs, the pairs on which any within-attempt signal must live.
The same probe, evaluated within
problem on those 22 problems ($n{=}5{,}632$) --- the within component of
Eq.~\eqref{eq:decomp}, estimated under both discordant-pair and the
pre-registered failure-count weighting (Appendix~\ref{app:extra}) ---
sits at $0.496$ $[0.466, 0.527]$ under the pre-registered
failure-count weighting and $0.515$ $[0.481, 0.562]$ under exact
discordant-pair weighting --- indistinguishable from chance under both
--- and remains so at every one of ten anchors from $t = 4$ to $512$ (length attrition
leaves 21 of the 22 problems with both outcome classes at $t = 512$;
Figure~\ref{fig:dissect}); the upper confidence limits cap the
failure-weighted and pair-weighted \emph{mean} within-problem AUROC at
$t = 4$ at $0.527$ and $0.562$ --- bounds on these weighted-mean
estimands, not on any single problem. The pooled positive is therefore predominantly
between-problem information, far below the
dump's full leave-one-out difficulty ceiling: scoring each attempt by the
pass rate of its problem's 255 \emph{other} attempts (the surrogate
$\hat d_{c,j}$ introduced with Eq.~\eqref{eq:decomp}, computed from the
full 32{,}768-sample verification table rather than the extracted subset)
and evaluating on the same pooled set yields AUROC $0.981$
$[0.955, 0.991]$. Moreover, with sample composition exactly fixed through
$t = 128$ (mild length attrition only at later anchors;
Appendix~\ref{app:extra}), the pooled AUROC itself collapses --- indistinguishable from chance by $t = 16$ ($0.586$,
CI straddling $0.5$) and $\approx 0.55$ from $t = 32$ onward: the
difficulty content readable from last-token states is a prompt echo that
fades as generation proceeds. This reconciles three observations in the literature ---
declining probe signal at larger reasoning budgets \citep{lugoloobi2026},
question-only probes that work before generation
\citep{cencerrado2025noanswer}, and the practice of probing at special
intermediate-\emph{answer} positions rather than arbitrary early positions
\citep{zhang2025knowright} --- under a single reading: what a linear probe reads from last-token
states is problem information that fades as generation proceeds, not the
fate of the attempt.

Two post-hoc checks, run after all frozen endpoints were reported
(Appendix~\ref{app:within-robust}), ask whether within-attempt information
could exist that a pooled-trained probe cannot see. A probe trained on
problem-centered states (each problem's mean removed, label-free) with
problem-disjoint folds is at chance at $t = 4$ (mean within-problem AUROC
$0.490$ $[0.464, 0.523]$) and reaches only $0.556$ $[0.522, 0.588]$ at
$t = 32$ and $0.52$--$0.54$ thereafter; a per-problem oracle finds the
signal concentrated in three of the 22 problems (AUROC $0.78$--$0.92$ at
$t = 4$, holding 84 of the 2{,}178 failures; above $0.74$ at every anchor
for two of the three, the third dipping to $0.60$ at $t = 256$), in two
of which the
failing samples are four to six times shorter than the successes --- an
early answer-without-reasoning mode. Within-attempt information therefore
exists but is small on average --- a transferable component appears only
from $t \approx 32$ --- and is concentrated where failures are rare, where
it is visible from the first tokens; the published-style positive draws
on none of it.

\section{Related Work}\label{sec:related}

\paragraph{Resampling probes of reasoning traces.}
Truncating a trace and sampling continuations has become the standard
counterfactual tool for interpreting chain-of-thought. \citet{lanham2023}
introduced early-answering truncation curves to measure how much the stated
reasoning matters; \citet{bogdan2025anchors}
resample sentence-level alternatives to attribute outcome changes to
individual steps; \citet{bigelow2025forking} resample at every token to
locate ``forking tokens'' where the outcome distribution shifts abruptly,
and \citet{bigelow2026forkingfast} show that much apparent per-token
volatility in such estimates is sampling noise. Closest in shape to our
probe, \citet{merrill2026ponr} fix prefixes and resample continuations at
scale to localize where a (deceptive) outcome ``locks in,'' and
\citet{ballon2026} read answer distributions from truncated prefixes,
finding that commitment rises steadily with prefix length; and
\citet{wang2026re2}, as motivation for training models to abandon bad
prefixes, compare continuing truncated \emph{incorrect} traces against
re-solving from scratch --- the nearest published continue-versus-restart
comparison, without matched token budgets or per-anchor value curves. All
of
these estimate what a prefix \emph{leads to}; none of them ask what the
prefix is \emph{worth} relative to not having it. Our contribution is the
missing control: an empty-prefix restart baseline $R(C)$ at matched total
generated-token budget, which converts per-anchor solve rates into a value measurement and
reveals that most apparent within-trace transitions are budget artifacts.

\paragraph{Rollout value estimates in process supervision.}
Our estimator --- the fraction of budgeted continuations from a partial
solution that reach the correct answer --- is the workhorse of automatic
process supervision: Math-Shepherd's soft label \citep{wang2024mathshepherd}
is exactly this quantity (its main experiments binarize it to whether any
continuation succeeds), OmegaPRM \citep{luo2024omegaprm} binary-searches
it for the first error, Phi-4's pivotal token search \citep{abdin2024phi4}
mines sharp jumps in it for preference pairs, and \citet{setlur2025progress}
formalize its increments as ``progress'' rewards. This literature treats the
quantity as training signal and its unreliability as label noise to be cleaned
\citep{zhang2025prmlessons}. We repurpose the same estimator as a
\emph{measurement instrument} --- with anchors on the model's own trace,
interval censoring, attempt-pooling, and the restart control --- and treat
its intermediate values as signal: roughly a third of ambiguous states
remain intermediate after eight attempts (success 0.3--0.6), which
undermines
the single-crossing, monotone-value picture implicit in binary-search
labeling schemes \citep{luo2024omegaprm} and in snowball-error theories of
monotone degradation \citep{gan2025snowball}.

\paragraph{Aha moments and their skeptics.}
The claim that reinforcement-trained reasoners exhibit emergent ``aha
moments'' originates in DeepSeek-R1's training anecdotes
\citep{deepseekr1}. Skepticism so far has been correlational or
lexical: reflection keywords exist before any RL \citep{liu2025noaha},
many reflection steps, self-verification among them, are causally inert \citep{zhao2025fakeaha,
kang2025firsttry}, and at trace scale, mid-reasoning strategy shifts are
rare and associated with \emph{lower} accuracy \citep{daliberti2026illusion}.
We supply the counterfactual, compute-matched version of this skepticism:
rather than classifying shifts in observed traces, we measure whether any
anchor of a trace carries value that fresh computation at equal total budget
cannot reproduce, and find such anchors in 1 of 178 cells. Sharp
within-trace transitions reported by entropy change-point analyses
\citep{xu2026entropy} and commitment-boundary probes \citep{scalena2026}
are consistent with our data --- but our restart control shows that the
states they mark are, almost always, reachable from scratch at the same
total cost.

\paragraph{Sequential versus parallel test-time compute.}
Aggregate comparisons of long chains against independent samples are now
common: revision models conditioned on complete prior answers
\citep{snell2024}, repeated-sampling coverage laws \citep{brown2024monkeys},
budget forcing \citep{muennighoff2025s1}, and comparisons at matched
token budgets \citep{sharma2025sequentialedge, ghosal2025mirage} or
matched sample counts \citep{pgap2026} finding either sequential
\citep{sharma2025sequentialedge} or parallel \citep{ghosal2025mirage,
pgap2026} advantages, with context contamination
as one proposed mechanism for failed retries \citep{yang2026retrying} and
reduced exploration when conditioning on prior answers \citep{pgap2026}. These comparisons operate at benchmark
level and condition on \emph{completed} answers. Our instrument moves the
comparison inside the trajectory --- the model's own truncated prefix
against a fresh start at matched total tokens, per problem --- and yields a
quantity the aggregate view cannot express: a prefix's value in fresh-token
units. The answer, for our models, is predominantly compression rather than
reachability (the prefix wins all nine exactly-matched comparisons, with
two wins and two ties on boundary proxies; an 8{,}192-token restart
reaches threshold in 11/13), which is empirical evidence for the no-benefit branch of the dichotomy of
\citet{wolf2026theory}: when self-reflection fails to reliably localize
early errors, conditioning on past attempts offers no asymptotic benefit
over independent restarts. Our per-problem restart dose--response curves also
give the sequential-parallel debate a diagnostic reading: one model's
failures dissolve with budget (compute-starved), the other's do not
(capability-limited), echoing overthinking \citep{chen2025overthinking}
and non-monotone thinking-length effects \citep{zhou2026morethinking}.

\paragraph{Predicting outcomes from internal signals.}
A rapidly growing line reports that hidden states predict reasoning
outcomes: probes at intermediate-answer positions reach AUROC $>0.9$
\citep{zhang2025knowright}, first-step probes 0.79
\citep{yuan2026diagnostic}, pre-generation probes beat question-text
baselines \citep{lugoloobi2026}, question-only probes predict success before
generation \citep{cencerrado2025noanswer}, and confidence signals drive
early exit and trace selection in systems work \citep{fu2024certaindex,
fu2025deepconf}. With few exceptions these positives
carry no problem-difficulty control, and the exceptions control only
partially: \citet{yuan2026diagnostic} run their within-problem analysis on
full-trace probes (their early-window positive is uncontrolled, and their
within-problem effect sizes fall as low as $d{=}0.13$ in their Table~3), while
\citet{lugoloobi2026} probe strictly before generation and themselves
conclude that activations encode model-specific \emph{difficulty}. Our
pre-registered, single-shot test supplies the missing control at the point
where it bites: frozen
summary features of the run's internal dynamics add no detectable AUROC
over a difficulty baseline on held-out problems at the pre-registered
forecast point, and a labeled post-hoc sweep finds no rescuing window
from 128 to 2{,}048 tokens --- consistent with reports that
correct and incorrect trajectories diverge late \citep{sun2026trajectories},
that trajectory geometry remains systematically coupled to difficulty once
length is residualized \citep{gjolbye2026}, and that difficulty-driven sample composition contaminates position-wise
probe curves, as \citet{david2025temporal} candidly note of their own
later-window decline --- their early-window positive carries no
question-only control, and it is the result we recover in a closely
matched reconstruction and then
reduce to chance within problem (Section~\ref{sec:results-prediction}). Existing critiques of correctness
probes attack two other axes: their signals resist causal use
\citep{yuan2026diagnostic}, and under contaminated prefixes they track
coherence rather than grounded correctness \citep{pair2026}. The
difficulty axis has remained open. Meanwhile, difficulty
itself is directly decodable from hidden representations
\citep{llmknowsdifficulty2025, geometrichardness2026}, which is precisely
the mechanism our dissection isolates.

\section{Discussion and Limitations}\label{sec:discussion}

\paragraph{What one cell in 178 means.}
We do not conclude that prefix-locked value is impossible --- our single
prefix-limited cell is an existence proof --- but that it is rare enough
in this regime that any account of reasoning built on frequent
within-trace unlocks is measuring budget fit. The practical corollary of
the compression finding is a reframing of restart policies: for a
compute-starved model, abandoning a long prefix typically costs a budget
multiple rather than access to the solution; for a capability-limited model,
neither continuation nor restart helps. Which regime a deployment faces is
measurable with a four-point restart curve.

\paragraph{What the prediction results do and do not show.}
Three layers --- a pre-registered null on our cohort, a trace-blind
difficulty ceiling inside the published range on 192K public R1
generations, and a
reconstruct-then-dissect analysis at 256 samples per problem --- support one
conclusion: pooled probe AUROCs on reasoning outcomes should be read as
difficulty measurements until a question-only baseline or within-problem
evaluation shows otherwise. The three layers carry unequal weight: the
dissection's bound on mean within-problem AUROC is the sharpest evidence;
the pre-registered cohort null is supporting, its primary interval
leaving moderate effects open ($+0.167$). This does not overturn any specific
published number. \citet{zhang2025knowright} probe intermediate-answer
positions and their data contain within-trace label variance; our claim
there is a falsifiable prediction: adding a question-only baseline and a
within-problem evaluation will substantially shrink the reported gaps. It
also does not make internal signals useless: within-problem trace
\emph{selection} \citep{fu2025deepconf} and pre-generation difficulty
routing \citep{lugoloobi2026} are consistent with everything we find ---
both exploit exactly the difficulty signal we isolate. Nor is
within-attempt information absent: a post-hoc within-targeted check
(Appendix~\ref{app:within-robust}) finds a component that is small on average and
concentrated in a few low-failure problems --- there the rare failures
are legible from the first tokens; the common failures on hard problems
are not --- which is not the signal the pooled positives report. Together with the
causal failures of probe signals \citep{yuan2026diagnostic} and their
coherence confound under contaminated prefixes \citep{pair2026}, the
critique now stands on three independent legs: probe signals can be causally
inert, coherence-driven, and difficulty-driven.

\paragraph{Limitations.}
Our instrumented cohort uses two small models in 4-bit quantization on one
benchmark family, with $m = 4$--$8$ attempts per cell and one base seed;
budget matching equates generated tokens, not FLOPs or latency
(continuing a stored prefix amortizes its KV cache, a restart recomputes
it, and attention cost grows with context --- none measured);
the confirmatory baseline is text-level (a pre-generation activation probe
would be stronger \citep{lugoloobi2026}); the public-dump dissection
re-scores full-precision generations through a 4-bit model (top-1 fidelity
0.906, reported) and inherits unknown sampling temperature; the probe
dissection itself is mathematics-only (the difficulty ceiling generalizes
off-math --- GPQA-diamond, Section~\ref{sec:results-prediction} --- but a
non-math dissection awaits a multi-sample dump from a locally runnable
model); and the
within-trajectory \emph{timing} question --- when does value arrive? ---
remained under-powered everywhere because genuine interior events are
rare, which is itself a finding: at this scale, breakthroughs are either
immediate or absent. No equivalence margin was pre-registered for the
confirmatory endpoints,
so their results are absences of detected gain rather than demonstrated
equivalence; and positive existence claims (the one prefix-limited cell,
the intermediate-solvability band) rest on modest counts and warrant
replication.

\paragraph{Future work.}
The natural next experiment runs the same frozen instrument on a
large RL-trained reasoner of the kind the aha-moment claims describe: interior
events there would revive the timing question with the controls already in
place, and their absence would extend the budget-artifact account to the
regime that motivated it. A second direction treats the value curve
$\phat(t)$ as the object of study --- its shape (gradual versus stepped),
not its crossings --- for which the present data already suffice.

\section*{Acknowledgements}
The external analyses in this paper required no new generation, because
others released their samples completely. We thank
Nishad Singhi and co-authors \citep{singhi2025solveverify,
singhi2025math128dump} for releasing per-problem solution dumps
that keep every attempt --- failures included --- at 256 samples per
problem; that design choice is what made the within-problem control of
Section~\ref{sec:results-prediction} possible. We likewise thank the
Hugging Face Open-R1 team for releasing OpenR1-Math-220k
\citep{openr1math} with per-generation correctness annotations rather
than a correct-only filter. We encourage the practice: unfiltered,
multi-sample rollout releases are reusable scientific instruments. This
study also builds on the MATH \citep{hendrycks2021math} and GPQA
\citep{rein2023gpqa} benchmarks and on open-weight models and tooling
(MLX and the mlx-community quantizations, scikit-learn, Hugging Face
datasets). AI assistance was used for coding and writing; all
experimental decisions, protocol freezes, and final claims were reviewed
by and are the responsibility of the author.

\bibliographystyle{plainnat}
\bibliography{references}

\appendix
\section{Protocol Amendments and Pre-Registration Timeline}\label{app:amendments}

Every decision rule in this study was frozen in a dated, committed
amendment before the outcomes it governs were observed. The full, dated
amendment documents ship with the code repository accompanying this
paper (\url{https://github.com/bulutyigit/problem-not-path});
this appendix condenses them in chronological order and records
how every gate resolved, including the failures.

\paragraph{Breakthrough forecasting protocol.}
Froze the base probe design on the development cohort: anchor grid,
$m = 4$ continuations, budget $B = 1{,}024 + 512$, threshold
$\tauthr = 0.75$ with next-anchor stability, interval censoring, bisection
refinement, the research splits (assigned before any outcome), and the
within-trajectory forecasting target
$P(\Tf \le t + k \mid \text{features through } t)$ whose power gate
Section~\ref{sec:results-prediction} later reports.

\paragraph{Probe sensitivity and supplement (A1--A3).}
Froze three responses to the development cohort's censoring pattern:
A1 terminal replication (threshold-censored cells gain four branches; a
terminal event requires pooled $\ge 6/8$); A2 budget sensitivity
(re-probing at a 4{,}096-token continuation budget with paired seeds); A3
a screened 20-problem supplement cohort. \emph{Resolution:} A2 falsified
the naive labels for Ministral-3 (16-token prefixes solve at the larger
budget) and validated Gemma-4's early crossings --- the observation that
forced the \Tf/\Tv{} distinction.

\paragraph{Restart-controlled breakthrough (A5).}
Froze the restart curves $R(C)$, the advantage definition at matched total
generated-token budget, $\Tv(\delta)$ with primary $\delta = 0.5$, the regime taxonomy
and its precedence order, and the conservative-interpolation sensitivity.
\emph{Pilot resolution:} the pilot gate required $8/8$ agreement and
scored $7/8$; recorded as a failure by the frozen wording, with the single
permitted enlargement resolving the borderline cell \emph{against} the
prior expectation.

\paragraph{Cohort expansion (waves and gates).}
The cohorts were collected in staged \emph{waves}; the expansion cohort
is wave~3, and released artifact files keep the wave-numbered names.
Froze the staged expansion: outcome-blind screens, wave gates with numeric
thresholds, and split assignment before generation. \emph{Gate
resolutions:} G1 (construct validity) failed for Ministral-3 per A2; G2
(wave-1 interior-event yield $\ge 6$) failed at 4, with the diagnosis
recorded --- the screen selected on 16K terminal solvability while probes
test 1{,}024-token continuations --- and the wave-3 screen was repaired
accordingly before any wave-3 outcome existed.

\paragraph{Ambiguity enlargement (A5.1).}
Froze the forward-looking rule that every $1$--$3/4$ probe cell is
enlarged to eight attempts before labels are derived, with thresholds
applied to pooled rates. \emph{Outcome:} labels were trimmed, not inflated
(six events removed, one added, project-wide).

\paragraph{Confirmatory early-signal test.}
Froze endpoints, feature sets, model class, folds, power gates, and the
success criterion; committed before the single test-split evaluation. The
same document's addendum labels the forecast-point sweep
($t \in \{128, 256, 1{,}024, 2{,}048\}$) as openly post-hoc.
\emph{Outcome:} both endpoints failed the success criterion; the horizon
endpoint stayed below its power gate and was not fit.

\paragraph{External re-analysis protocol (reconstruction).}
Froze a faithful reconstruction of the closest published early-window
positive (model, data construction, probe recipe) plus the added controls.
Superseded before execution by the public-dump design below, which
dominates it (more data, no generation, and a within-problem control);
the apparatus remains in the repository.

\paragraph{Public-dump probe dissection.}
Froze, after sample verification but before any state extraction: the
trajectory selection (all 256 samples of the 22 mid-band problems plus 8
per problem elsewhere), the teacher-forcing extraction, the probe recipe,
both evaluation metrics with their reference rows, and the reading key for
either outcome. \emph{Outcome:} recorded in the same document; reported in
Section~\ref{sec:results-prediction}. A within-problem robustness check
run after reporting is labeled post-hoc in the same document and in
Appendix~\ref{app:within-robust}.

\section{Additional Figures and Sensitivity Analyses}\label{app:extra}

\subsection{Early-window signal definitions}\label{app:signals}

For each generated token $t$ let $p_t$ be the model's next-token
distribution over vocabulary $V$ at the position that emits token $x_t$,
and $h_t \in \mathbb{R}^d$ the final-layer hidden state at that position.
The instrumented token-level series are:
\begin{align*}
\tilde H_t &= -\frac{1}{\log \lvert V\rvert} \sum_{v \in V} p_t(v)\, \log p_t(v)
  && \text{normalized entropy}\\[3pt]
u_t &= -\log p_t(x_t)
  && \text{surprisal}\\[3pt]
\mu_t &= p_t^{(1)} - p_t^{(2)}
  && \text{top-1/top-2 probability margin}\\[3pt]
\mathrm{KL}_t &= \mathrm{KL}\bigl(p_t \,\big\|\, p_{t-1}\bigr)
  && \text{successive KL divergence}\\[3pt]
\mathrm{JS}_t &= \tfrac{1}{2}\,\mathrm{KL}\bigl(p_t \,\big\|\, m_t\bigr)
  + \tfrac{1}{2}\,\mathrm{KL}\bigl(p_{t-1} \,\big\|\, m_t\bigr),
  \quad m_t = \tfrac{1}{2}\bigl(p_t + p_{t-1}\bigr)
  && \text{successive JS divergence}\\[3pt]
\nu_t &= \lVert h_t \rVert
  && \text{hidden-state norm}\\[3pt]
\rho_t &= \lVert h_t - h_{t-1} \rVert \big/ \lVert h_{t-1} \rVert
  && \text{relative hidden step}\\[3pt]
\gamma_t &= 1 - \cos\bigl(h_t,\, h_{t-1}\bigr)
  && \text{cosine drift}
\end{align*}
Over the forecast window ($t \le 512$), each series is reduced by simple
statistics: the mean and standard deviation; the \emph{robust slope}, a
Theil--Sen estimator (median of pairwise slopes, evaluated on at most 512
evenly spaced points for bounded cost); and for surprisal the maximum
one-step rise $\max_t (u_{t+1} - u_t)$. Spectral summaries are computed on
the $\mathrm{JS}_t$ series: robust-standardize by median and median
absolute deviation, resample to 256 points, apply a Hann window, take the
real FFT power spectrum with the DC term removed, and normalize to
$\pi_f$; then
\[
\mathrm{SpecEnt} = -\tfrac{1}{\log F} \textstyle\sum_f \pi_f \log \pi_f,
\qquad
\mathrm{Centroid} = \textstyle\sum_f f\, \pi_f,
\qquad
\mathrm{LowRatio} = \textstyle\sum_{f \le 0.10} \pi_f .
\]
{\sloppy The fifteen frozen summaries of Section~\ref{sec:results-prediction} are:
$\mathrm{mean}(\tilde H)$, $\mathrm{sd}(\tilde H)$,
$\mathrm{slope}(\tilde H)$; $\mathrm{mean}(u)$, $\max$-rise$(u)$;
$\mathrm{mean}(\mu)$; $\mathrm{mean}(\mathrm{JS})$,
$\mathrm{sd}(\mathrm{JS})$, $\mathrm{slope}(\mathrm{KL})$;
$\mathrm{mean}(\rho)$, $\mathrm{mean}(\gamma)$, $\mathrm{slope}(\nu)$;
and $\mathrm{SpecEnt}$, $\mathrm{Centroid}$, $\mathrm{LowRatio}$ of
$\mathrm{JS}_t$. The public-dump probes of
Section~\ref{sec:results-prediction} reuse the unnormalized entropy,
$u_t$, and top-1 probability series from teacher-forced forwards (the
normalization constant is fixed within a model and immaterial after
standardization).\par}

\subsection{The budget falsification that motivated the restart control}

\begin{figure}[h]\centering
\includegraphics[width=\linewidth]{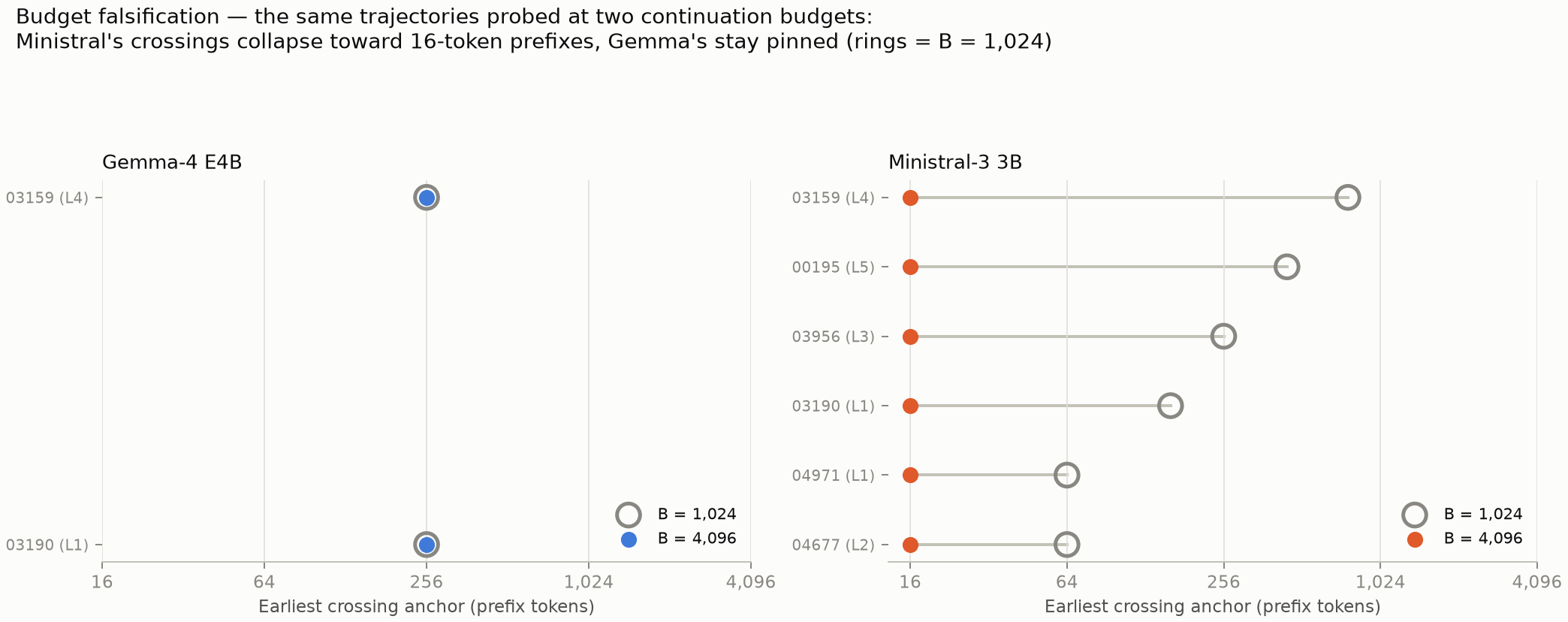}
\caption{Budget sensitivity (amendment A2,
App.~\ref{app:amendments}): the same cells probed at $B = 1{,}024$
versus $B = 4{,}096$. Ministral-3's ``breakthroughs'' are budget
artifacts; Gemma-4's survive.}
\label{fig:budgetfals}\end{figure}

Figure~\ref{fig:budgetfals} shows the A2 experiment behind
Section~\ref{sec:results-regimes}: development-cohort cells re-probed at a
4{,}096-token continuation budget with paired seeds. Ministral-3's
apparent mid-trajectory crossings dissolve --- 16-token prefixes solve at
the larger budget --- while Gemma-4's early crossings remain pinned,
validating them as genuine.

\subsection{Per-trajectory intervals and label hardening}

\begin{figure}[htbp]\centering
\includegraphics[width=\linewidth]{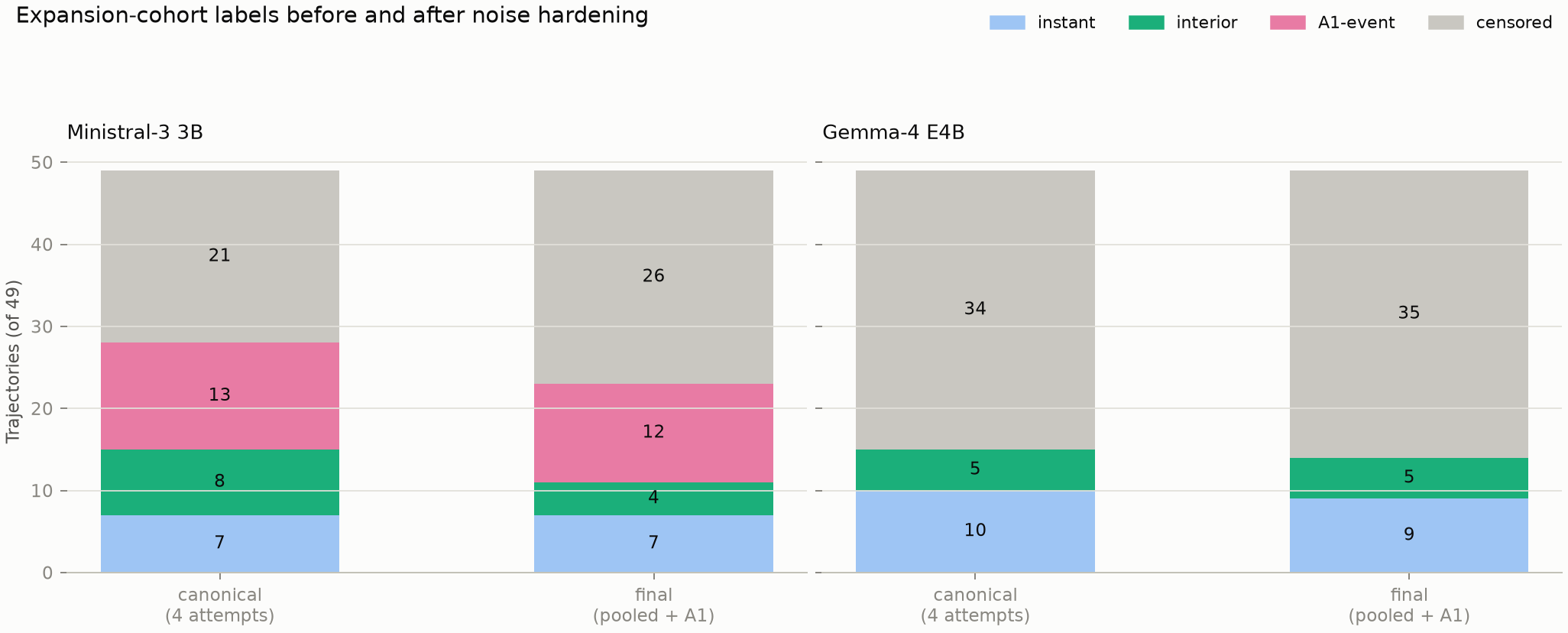}
\caption{Expansion-cohort labels before and after noise hardening
(pooling + terminal replication): trimmed, not inflated.}
\label{fig:hardening}\end{figure}

\begin{figure}[p]\centering
\includegraphics[width=\linewidth]{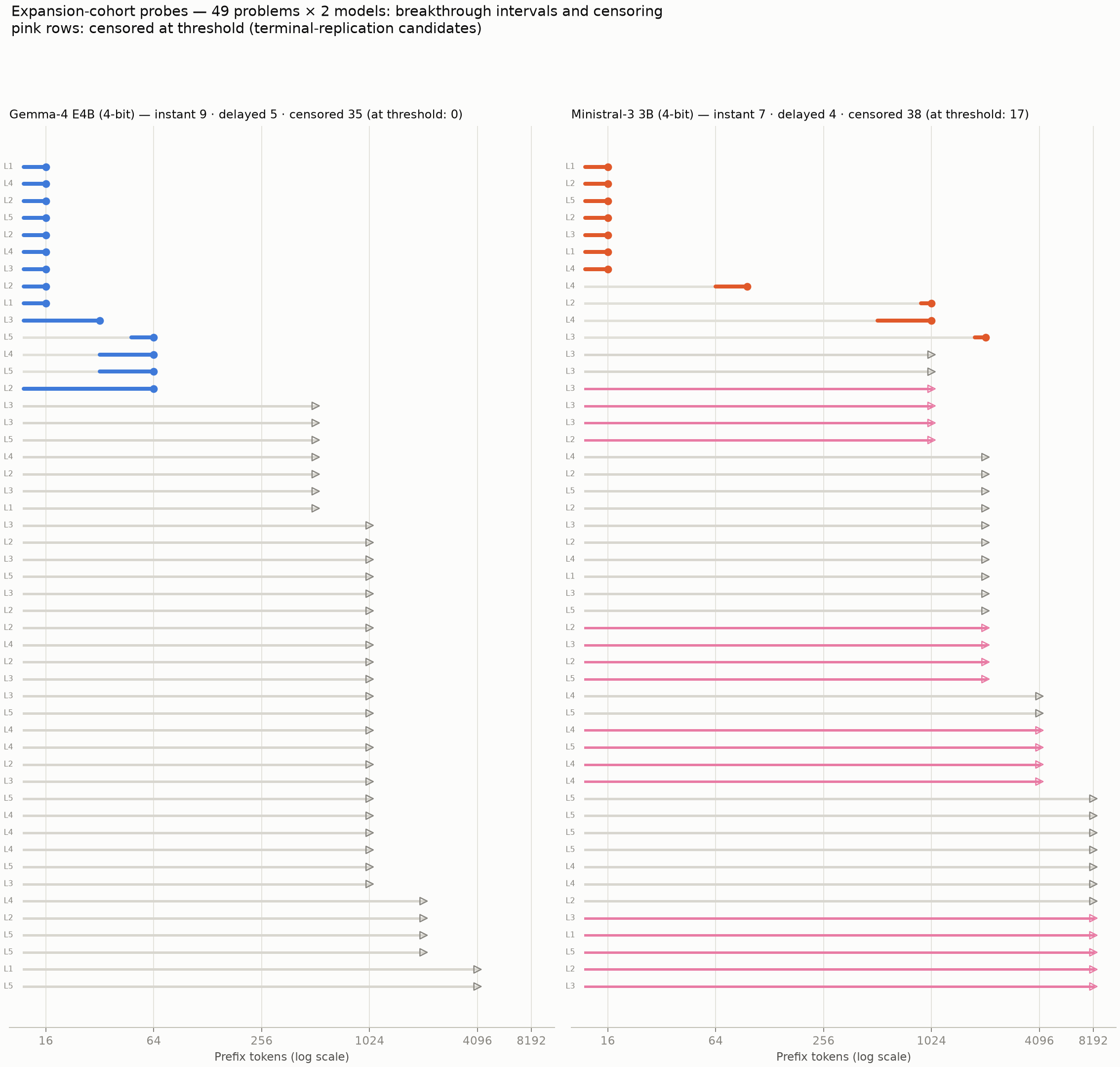}
\caption{Expansion cohort, one row per trajectory: breakthrough intervals
(solid), censoring bounds (arrows), and threshold-censored replication
candidates (highlighted).}
\label{fig:swimmer}\end{figure}

Figure~\ref{fig:hardening} shows what the noise-hardening rules changed;
Figure~\ref{fig:swimmer} shows the breakthrough interval or censoring
bound of every expansion-cohort trajectory. Terminal replication confirmed 13
of Ministral-3's 16 threshold-censored candidates and rejected Gemma-4's
single candidate; pooling ambiguous cells to eight attempts moved labels
on $7+6$ of $49+49$ trajectories, dominated by losses (six events removed,
one added; Gemma-4's changes were mostly interval shifts).

\subsection{Sensitivity of \Tv{} labels}

Independent-half check for the intermediate band: on branches 4--7 alone
--- never used for selection --- the 148 enlarged cells land at
$0/1/2/3/4$ successes in $37/37/28/21/25$ cells; 58\% fall in the
intermediate $1$--$3$ range, against the concentration at the extremes
that a pure selection artifact would predict.

Across all 178 cells: $\delta = 0.5$ yields 3 prefix-value events,
$\delta = 0.25$ yields 7 (the additions all sit inside budget-limited
cells of the budget-elastic model), $\delta = 0.75$ keeps 2. The
conservative variant of $\Rhat$ (upper envelope of the bracketing grid
values) agrees with the primary labels on 178 of 178 cells. Cross-model
agreement on collapsed regime classes is 42 of 89 problems.

\subsection{Confirmatory test: level-free variants and per-model detail}

\begin{table}[h]\centering\small
\begin{tabular}{llccc}
\toprule
Endpoint & Variant & Baseline & +Early signals & $\Delta$AUROC [95\% CI] \\
\midrule
Primary   & main       & 0.828 & 0.853 & $+0.026$ $[-0.054, +0.167]$ \\
Primary   & level-free & 0.914 & 0.879 & $-0.035$ $[-0.156, +0.024]$ \\
Secondary & main       & 0.875 & 0.785 & $-0.090$ $[-0.213, +0.033]$ \\
Secondary & level-free & 0.840 & 0.771 & $-0.070$ $[-0.174, +0.056]$ \\
\bottomrule
\end{tabular}
\caption{Held-out test AUROCs for both endpoints, with the difficulty
level dropped from both feature sets (level-free). No variant meets the
success criterion.}
\label{tab:levelfree}\end{table}

Per-model test-split descriptives (no per-model claims are made): on the
primary endpoint the baseline scores 1.00 for Gemma-4 ($n{=}16$) and 0.69
for Ministral-3 ($n{=}17$, early signals 0.76); on the secondary endpoint
1.00 versus 0.73 ($n{=}11$, $14$). The horizon endpoint's power
arithmetic: 21 interior events project-wide, 4 in the test split; the
frozen gate (five problem groups per class per fold) was not met and no
model was fit.

\subsection{Public-dump dissection: full anchor table}

\begin{table}[htbp]\centering\small
\begin{tabular}{rcccc}
\toprule
$t$ & Pooled AUROC [95\% CI] & Within (failure-w.) [95\% CI] & Within (pair-w.) [95\% CI] & rows \\
\midrule
4   & 0.849 $[0.735{,} 0.919]$ & 0.496 $[0.466{,} 0.527]$ & 0.515 $[0.481{,} 0.562]$ & 6{,}480 \\
8   & 0.678 $[0.503{,} 0.835]$ & 0.481 $[0.461{,} 0.502]$ & 0.498 $[0.480{,} 0.515]$ & 6{,}480 \\
16   & 0.586 $[0.461{,} 0.706]$ & 0.495 $[0.467{,} 0.526]$ & 0.510 $[0.478{,} 0.552]$ & 6{,}480 \\
32   & 0.561 $[0.422{,} 0.702]$ & 0.525 $[0.494{,} 0.555]$ & 0.526 $[0.490{,} 0.568]$ & 6{,}480 \\
64   & 0.582 $[0.442{,} 0.706]$ & 0.481 $[0.451{,} 0.510]$ & 0.478 $[0.452{,} 0.508]$ & 6{,}480 \\
128   & 0.548 $[0.425{,} 0.664]$ & 0.515 $[0.491{,} 0.542]$ & 0.509 $[0.485{,} 0.538]$ & 6{,}480 \\
192   & 0.572 $[0.421{,} 0.706]$ & 0.499 $[0.466{,} 0.525]$ & 0.499 $[0.475{,} 0.527]$ & 6{,}478 \\
256   & 0.565 $[0.406{,} 0.684]$ & 0.530 $[0.476{,} 0.591]$ & 0.509 $[0.483{,} 0.543]$ & 6{,}476 \\
384   & 0.547 $[0.432{,} 0.650]$ & 0.509 $[0.482{,} 0.537]$ & 0.513 $[0.494{,} 0.536]$ & 6{,}294 \\
512   & 0.547 $[0.428{,} 0.656]$ & 0.497 $[0.468{,} 0.526]$ & 0.499 $[0.477{,} 0.521]$ & 6{,}102 \\
\bottomrule
\end{tabular}
\caption{Hidden-state probe on public R1-Distill-7B trajectories at every
anchor. The rows column counts trajectories entering the problem-disjoint
out-of-fold prediction; the \emph{pooled} metric is evaluated on the
8-per-problem pooled subset ($n{=}1{,}024$ at $t{=}4$, shrinking with
window attrition) and the \emph{within} metrics on the 22 mid-band
problems ($n{=}5{,}632$; 2{,}178 failures; at $t{=}512$ attrition leaves
21 problems with both outcome classes and 2{,}114 failures), under the pre-registered
failure-count weighting and the exact discordant-pair weighting of
Eq.~\eqref{eq:decomp}. References at $t{=}4$: full-dump LOO pass-rate
pooled AUROC 0.981 $[0.955{,} 0.991]$ (255 other attempts per problem,
evaluated on the same pooled set); stream-summary probe
pooled 0.647, within 0.490. The within-AUROC of the LOO score itself is
mechanically 0 (within a problem it takes two values, and the incorrect
attempt's is always higher), so the neutral reference for ``no
within-attempt information'' is 0.5.}
\label{tab:dissect}\end{table}

Numerical robustness: the float32 battery reproduces in float64 with zero
NaN out-of-fold predictions; scikit-learn overflow warnings trace to
near-constant standardized dimensions and are cosmetic. Teacher-forcing
fidelity: mean top-1 agreement 0.906 (10th percentile 0.859).

\subsection{Post-hoc within-problem robustness checks}\label{app:within-robust}

Run after all frozen endpoints were reported and labeled post-hoc in the
amendment, these checks ask whether within-attempt information could
exist that the pooled-trained probe of
Section~\ref{sec:results-prediction} cannot see. Both use the stored
states of the 22-problem within set ($n{=}5{,}632$). \textbf{A,
problem-centered transfer:} each problem's mean state is subtracted
(label-free) and the probe (standardization, PCA${}\le 128$, balanced
logistic) is trained on problem-disjoint folds and scored within held-out
problems. \textbf{B, per-problem oracle:} a separate probe (PCA 32) per
problem with sample-disjoint folds --- an upper bound on linearly
decodable within-problem signal that requires no transfer. An uncentered
re-run on the within set alone reproduces the frozen battery ($0.515$
$[0.488, 0.547]$ at $t{=}4$ against $0.496$ $[0.466, 0.527]$, which trains
on both sets). Table~\ref{tab:withinrobust} reports failure-weighted means
with problem-clustered bootstrap CIs.

The signal is heterogeneous. Three problems (ids 20, 22, and 13, with 18,
49, and 17 failures --- 84 of 2{,}178) are strongly separable at $t{=}4$ (oracle AUROC $0.92$, $0.84$, $0.78$;
centered transfer $0.93$, $0.86$, $0.33$); across all ten anchors the
oracle stays above $0.74$ for problems 20 and 22, while problem 13 dips to
$0.60$ at $t{=}256$ and loses its failures to attrition by $t{=}512$;
per-problem oracle AUROCs are consistent across anchors (Spearman $0.71$ between $t{=}32$ and $64$). In
problems 20 and 22 the failing samples are four to six times shorter than
the successes (median 2{,}364 versus 15{,}321 and 1{,}299 versus 5{,}168
characters): an early answer-without-reasoning mode, visible from the
first tokens. In problem 13 lengths and re-scoring fidelity match --- a
genuine early content divergence. The high-failure problems that dominate
the failure-weighted mean (216 and 223 failures of 256, for instance)
show no separability. Within-attempt information therefore exists but is
small on average (a transferable component appears only from
$t \approx 32$) and is concentrated where failures are rare, where it is
visible from the first tokens --- the opposite of the regime that produces the
pooled positive.

\begin{table}[h]\centering\small
\begin{tabular}{rcccc}
\toprule
$t$ & A: centered transfer [95\% CI] & B: per-problem oracle [95\% CI] & B max & B $>0.6$ \\
\midrule
4   & 0.490 $[0.464{,} 0.523]$ & 0.476 $[0.435{,} 0.529]$ & 0.924 & 3 \\
8   & 0.526 $[0.502{,} 0.550]$ & 0.479 $[0.449{,} 0.519]$ & 0.873 & 3 \\
16  & 0.486 $[0.451{,} 0.534]$ & 0.481 $[0.437{,} 0.536]$ & 0.908 & 4 \\
32  & 0.556 $[0.522{,} 0.588]$ & 0.519 $[0.478{,} 0.560]$ & 0.894 & 5 \\
64  & 0.532 $[0.502{,} 0.565]$ & 0.558 $[0.523{,} 0.593]$ & 0.868 & 7 \\
128 & 0.524 $[0.504{,} 0.548]$ & 0.519 $[0.489{,} 0.556]$ & 0.824 & 4 \\
192 & 0.526 $[0.498{,} 0.556]$ & 0.491 $[0.452{,} 0.543]$ & 0.900 & 5 \\
256 & 0.536 $[0.503{,} 0.570]$ & 0.531 $[0.493{,} 0.576]$ & 0.935 & 5 \\
384 & 0.529 $[0.516{,} 0.546]$ & 0.532 $[0.493{,} 0.576]$ & 0.939 & 6 \\
512 & 0.521 $[0.466{,} 0.580]$ & 0.507 $[0.457{,} 0.549]$ & 0.932 & 3 \\
\bottomrule
\end{tabular}
\caption{Post-hoc within-problem checks on the 22-problem within set:
failure-weighted mean within-problem AUROC of a problem-centered transfer
probe (A) and of per-problem oracles (B), the largest single-problem
oracle AUROC, and the number of problems above 0.6. Within-problem
label-permutation nulls (12 draws at $t \in \{4, 32, 64, 128\}$) span
$0.47$--$0.54$ for the A means, $0.44$--$0.54$ for the B means, and
$0.53$--$0.76$ for the B maxima.}
\label{tab:withinrobust}\end{table}

\subsection{Post-hoc forecast-point sweep: full table}

Table~\ref{tab:sweep} lists both endpoints at every forecast point.

\begin{table}[h]\centering\small
\begin{tabular}{rcccc}
\toprule
$t$ & Primary $\Delta$ [95\% CI] & $n$ & Secondary $\Delta$ [95\% CI] & $n$ \\
\midrule
128    & $-0.083$ $[-0.281, +0.016]$ & 34 & $-0.118$ $[-0.274, +0.048]$ & 25 \\
256    & $+0.092$ $[-0.032, +0.384]$ & 34 & $-0.035$ $[-0.133, +0.053]$ & 25 \\
512*   & $+0.026$ $[-0.054, +0.167]$ & 33 & $-0.090$ $[-0.213, +0.033]$ & 25 \\
1{,}024 & $-0.125$ $[-0.409, 0.000]$ & 24 & $-0.137$ $[-0.287, +0.018]$ & 22 \\
2{,}048 & $-0.133$ $[-0.364, 0.000]$ & 13 & $-0.259$ $[-0.800, 0.000]$ & 12 \\
\bottomrule
\end{tabular}
\caption{$\Delta$AUROC (early signals $-$ baseline) at every forecast
point; * marks the pre-registered evaluation, all others post-hoc. Larger
prefixes also shrink and select the sample (only runs surviving to $t$
enter).}
\label{tab:sweep}\end{table}

\subsection{Reproducibility details}\label{app:repro}

\textbf{Generation.} Both models sample at temperature $0.6$, top-$p$
$0.95$, top-$k$ $20$ (no repetition or presence penalties), thinking mode
enabled, prompt version v1 (a fixed task instruction requesting a final
boxed answer), 16{,}384-token cap for base trajectories. Probe branches
use deterministic seeds derived as
$\mathrm{sha256}(\text{run},\text{anchor},\text{branch})$; restart
attempts hash the problem, model, budget, and branch. Model checkpoints
are pinned to the revisions recorded in the released readiness manifests
(\texttt{mlx-community} 4-bit conversions of the checkpoints named in
Section~\ref{sec:method}).

\textbf{Verification.} A final answer is extracted from the generated
text; extraction failure scores as incorrect. Extracted answers are
compared to the reference by numeric equivalence first, then symbolic
equivalence. GPQA answers compare the extracted option letter
(``final answer is X'') against the key; unparseable samples ($3.0\%$)
score incorrect.

\textbf{Cohort screens.} Development and supplement problems were drawn
from a 100-problem level-balanced pool by frozen outcome-blind rules; the
expansion screen selected problems with 1--5 verified successes out of 6
fresh 3{,}072-token attempts (2 models $\times$ 3 seeds).

\textbf{Classifiers.} The instrumented-cohort analyses (confirmatory
test and sweep) use $\ell_2$ logistic regression with C $= 0.1$,
liblinear, class-weight balanced, at most 5{,}000 iterations, on
standardized features. The public-dump probe battery uses scikit-learn
defaults (C $= 1$, lbfgs) with class-weight balanced, at most 2{,}000
iterations, standardization, and PCA to at most 128 components ---
matching the probe recipes it reconstructs. Folds are StratifiedGroupKFold
with problems as groups throughout. Confidence intervals are percentile
bootstraps over 2{,}000 problem-clustered draws (1{,}000 for the post-hoc
checks of Appendix~\ref{app:within-robust}).

\textbf{Exclusions.} Fifteen of 178 cells whose base run ended before
the 512-token forecast window (10 train, 4 validation, 1 test) are
excluded from the confirmatory endpoints, per the frozen inclusion rule.

\end{document}